\documentclass[letterpaper, 10 pt, conference]{ieeeconf}
\IEEEoverridecommandlockouts
\usepackage{setspace}
\usepackage{amsmath}
\usepackage{amsfonts}
\usepackage{cite}
\usepackage{url}
\usepackage{ragged2e}
\usepackage{soul}

\usepackage{booktabs} 
\usepackage{multirow}
\usepackage{diagbox}
\usepackage{tabularx}
\usepackage{makecell}
\usepackage{float}

\usepackage{graphicx}
\usepackage{tikz}
\usepackage[font=footnotesize]{caption}
\usepackage{subcaption}

\usepackage[ruled,vlined, linesnumbered, noend]{algorithm2e}
\usepackage{optidef}

\usepackage[bookmarks=false, hidelinks]{hyperref}
\usepackage{cleveref}

\newtheorem{remark}{Remark}
\newtheorem{problem}{\bf{Problem}}

\title{\bf Energy-Aware Collaborative Exploration for a UAV–UGV Team}

\author{Cahit Ikbal Er, Saikiran Juttu, and Yasin Yaz{\i}c{\i}o\u{g}lu \thanks{C. I. Er and S. Juttu are with the Department of Mechanical and Industrial Engineering at Northeastern University, Boston, MA.}
\thanks{Y. Yaz{\i}c{\i}o\u{g}lu is with the Departments of Mechanical and Industrial Engineering and Electrical and Computer Engineering at Northeastern University, Boston, MA.  }\thanks {e-mails: \{er.c, juttu.s, y.yazicioglu\}@northeastern.edu.}}

\begin{document}
\bstctlcite{bstctl:etal, bstctl:nodash, bstctl:simpurl}
\maketitle
\thispagestyle{empty}
\pagestyle{empty}

\justifying

\begin{abstract}
We present an energy-aware collaborative exploration framework for a UAV–UGV team operating in unknown environments, where the UAV’s energy constraint is modeled as a maximum flight-time limit. The UAV executes a sequence of energy-bounded exploration tours, while the UGV simultaneously explores on the ground and serves as a mobile charging station. Rendezvous is enforced under a shared time budget so that the vehicles meet at the end of each tour before the UAV reaches its flight-time limit. We construct a sparsely coupled air–ground roadmap using a density-aware layered probabilistic roadmap (PRM) and formulate tour selection over the roadmap as coupled orienteering problems (OPs) to maximize information gain subject to the rendezvous constraint. The resulting tours are constructed over collision-validated roadmap edges. We validate our method through simulation studies, benchmark comparisons, and real-world experiments.
\end{abstract}

\section{Introduction}
Robotic exploration aims to autonomously expand a robot’s knowledge of unknown environments. Classical approaches rely on frontier-based strategies \cite{yamauchi1997frontier}, while information-driven methods such as next-best-view (NBV) and receding-horizon planning improve efficiency by reasoning about expected information gain \cite{bircher2016rhnbv}. Hybrid and sampling-based approaches, including incremental probabilistic roadmap (PRM) methods \cite{selin2019efficient, xu2023heuristic}, further balance coverage and computational tractability. Exploration has also been extended to multi-robot systems through coordinated frontier representations \cite{soni2022multi, kit2025race}, graph-based planning in complex environments \cite{dang2020graph}, coordination under limited communication \cite{bramblett2022multi}, and orienteering-based routing for heterogeneous teams with independent per-robot budgets \cite{sakamoto2020routing}.

Among heterogeneous teams, UAV–UGV systems are particularly attractive for monitoring and exploration missions because of their complementary capabilities. UAVs provide agile aerial sensing and rapid viewpoint changes, while UGVs offer long-endurance ground mobility and greater payload capacity. Several works have studied energy-aware planning of such systems, where the UGV acts as a mobile charging station rather than as an active contributor to the monitoring or exploration task \cite{ropero2019terra, yu2018algorithms, maini2019coverage, Lin2022, er2025rspect}. Despite extensive work on exploration and UAV–UGV coordination, their integration as collaborative co-explorers that explicitly account for energy and recharging constraints remains limited. In \cite{patel2025hierarchical}, a hierarchical graph-based exploration framework is introduced, where the UGV selectively deploys a UAV based on confidence-driven frontier assessment and expected gain. In \cite{depetris2022marsupial} and \cite{linq2022multimodal}, marsupial legged-aerial systems are presented, where the aerial robot is deployed only when the ground robot encounters inaccessible terrain. In \cite{delmerico2017active}, the UAV explores solely to map terrain for UGV path planning, while in \cite{zheng2025aage}, the UAV provides bird's-eye view to guide UGV exploration in large-scale environments. 

In a related work \cite{butzke2015airgnd}, coordinated 3-D exploration by a UAV-UGV team is performed using a centralized map-merging and goal-assignment framework. While both robots contribute to exploration, the UAV’s energy limitation is handled operationally (e.g., forced landing when the battery is low) rather than embedded as a structural constraint within the exploration optimization. Goals are assigned without explicitly enforcing energy-feasible rendezvous or modeling repeated recharging cycles. Consequently, exploration is not formulated under an explicit energy-constrained rendezvous framework required for long-horizon missions.


In this paper, we propose an energy-aware collaborative exploration framework for a UAV–UGV team in which both robots actively explore using tours planned under a shared time budget derived from the UAV's flight-time constraint. Between consecutive tours, the UAV is recharged by the UGV. The key challenge is coordinating both robots so that exploration remains energy-feasible while maximizing information gain. For planning the exploration tours, we maintain a density-aware layered probabilistic roadmap (PRM) that sparsely couples aerial and ground configurations via rendezvous connections. We formulate tour selection as coupled orienteering problems (OPs) on the PRM: an aerial OP and a ground OP, each maximizing their robot's information gain. These OPs are coupled through a shared pair of start and end points that embeds the rendezvous and the flight-time constraint imposed on both tours simultaneously. The main contributions of this work are as follows:

\begin{enumerate}
\item A joint formulation of flight-time–constrained collaborative exploration for UAV–UGV teams that explicitly embeds a shared rendezvous mechanism into long-horizon exploration planning (Problem~\ref{prob:exploration}).

\item A coordinated planning framework that combines a sparsely coupled layered PRM with coupled orienteering problems (OPs) to generate flight-time–feasible, concurrent UAV–UGV exploration tours that maximize joint information gain.

\item Extensive validation through simulation studies, benchmark comparisons, and real-world experiments.
\end{enumerate}

\section{Problem Formulation}
\label{sec:problem_formulation}

We consider an exploration mission, where an energy-aware UAV-UGV team is tasked with exploring an initially unmapped (unknown) environment. Both the UAV and the UGV actively explore the environment, and the UGV additionally serves as a mobile charging station, enabling the UAV to recharge between exploration tours. We denote the environment of the mission as $\mathcal{V}$, which contains the free space $\mathcal{V}_\text{free} \subset \mathcal{V}$, and occupied space $\mathcal{V}_\text{occ} \subset \mathcal{V}$. Due to constraints such as robot size, sensor range, and environment geometry, etc. there exists some residual (unreachable) space, $\mathcal{V}_\text{res} \subset \mathcal{V}$. For representing the environment, an occupancy map $\mathcal{M}$ will be employed, dividing the space $\mathcal{V}$ into cubical voxels with resolution $r$. These cubical voxels, $m\in\mathcal{M}$, can be marked as free, occupied, or unmapped. We denote $\mathcal{V}_{\text{mapped}} \subset \mathcal{V}$ as the subset of the environment whose corresponding voxels in $\mathcal{M}$ have been classified as either free or occupied. Initially, the whole environment is unmapped, except for the nearby region of the UAV-UGV team, $\mathcal{V}^0_{\text{mapped}}$. 

Let $\mathcal{C}_{\text{uav}} := \mathcal{V} \times S^1$ denote the configuration space for the UAV, where $S^1$ represents the yaw angle. Similarly, let $\mathcal{C}_{\text{ugv}} := \mathcal{V}_{\text{ugv}} \times S^1$ denote the configuration space for the UGV, where $\mathcal{V}_{\text{ugv}} \subset \mathcal{V}$ is the ground surface. The sets of feasible (collision-free) configurations for the robots are then denoted by $\Xi_{\text{uav}} \subset \mathcal{C}_{\text{uav}}$ and $\Xi_{\text{ugv}} \subset \mathcal{C}_{\text{ugv}}$, respectively. The UAV and the UGV can rendezvous at feasible configurations satisfying $\xi \in \Xi_{\text{uav}} \cap \Xi_{\text{ugv}}$. 

During exploration, the UAV should never be expected to continuously fly longer than a specified duration, $\overline{\tau_a}$, due to its energy limitations. Here, $\overline{\tau_a}$ represents the maximum flight time, i.e., available time excluding takeoff/landing operations. In order to satisfy this constraint, we adopt a strategy where the UAV performs a series of energy-constrained exploration tours. Each tour involves a bounded-duration flight segment, including: UAV taking-off from the UGV, exploring a part of the environment, and landing on the UGV for recharging in a potentially different position. Between tours, the UAV remains on the UGV for a duration $\tau_c$ to allow battery swap/charging such that the UAV is fully charged before the next tour. We denote the take-off positions as \textit{release points} and landing positions as \textit{collect points}, both of which are \textit{rendezvous points}. During UAV's exploration tour, the UGV also explores the environment and meets with the UAV at \textit{collect} points. We define the problem as:

\begin{problem}[Energy-Aware Collaborative Exploration]
\label{prob:exploration}
Given $\mathcal{V}_{\text{mapped}}^0$,  the objective of each tour is to maximize the joint information gain of both robots subject to the energy and rendezvous constraints below, thereby incrementally expanding $\mathcal{V}_{\text{mapped}}$. In each tour, the energy-limited UAV needs to follow a feasible path (sequence of configurations) $\sigma_{\text{uav}} = (\xi_1^{\text{uav}}, \xi_2^{\text{uav}}, \dots, \xi_K^{\text{uav}})$, $\xi_i^{\text{uav}} \in \Xi_{\text{uav}}$ for all $i$, whose flight time satisfies $\tau_a(\sigma_{\text{uav}}) \leq \overline{\tau_a}$, where $\tau_a(\cdot)$ is the travel time of a path for the UAV. Concurrently, the UGV must follow its own feasible path (sequence of configurations) $\sigma_{\text{ugv}} = (\xi_1^{\text{ugv}}, \xi_2^{\text{ugv}}, \dots, \xi_M^{\text{ugv}})$, $\xi_j^{\text{ugv}} \in \Xi_{\text{ugv}}$ for all $j$, and must reach the collect point within $\overline{\tau_a}$ to rendezvous with the UAV, i.e., $\tau_g(\sigma_{ugv})\leq\overline{\tau_a}$ where $\tau_g(\cdot)$ is the travel time of a path for the UGV. These paths must be coordinated such that the UGV starts and ends each tour at the projected configurations of the UAV’s release and collect points, i.e., $\xi_1^{\mathrm{ugv}} = \pi(\xi_1^{\mathrm{uav}})$ and $\xi_{M}^{\mathrm{ugv}} = \pi(\xi_{K}^{\mathrm{uav}})$. Here, $\pi : \mathcal{C}_{\mathrm{uav}} \rightarrow \mathcal{C}_{\mathrm{ugv}}$ is a projection function that maps an aerial configuration to its nearest feasible ground configuration. By executing these tours, the team jointly and incrementally expands the mapped region $\mathcal{V}_{\text{mapped}}$ through their onboard sensing while the UGV has the additional role of recharging the UAV at the end of each tour. The number of tours is not fixed a priori; the mission continues until the team explores all reachable space, i.e.,
\begin{equation*}
    \mathcal{V}_{\text{mapped}} = (\mathcal{V}_{\text{free}} \cup \mathcal{V}_{\text{occ}}) \setminus \mathcal{V}_{\text{res}}.
\end{equation*}
\end{problem}

\section{Proposed Method}
\label{sec:proposed_method}
Our approach utilizes a layered roadmap and a coupled planning framework to coordinate the motions of the UAV and the UGV during exploration. First, we construct a layered probabilistic roadmap (PRM) by sampling aerial configuration space and generating a set of feasible ground configurations that include both projections of aerial nodes when possible and additional independently sampled ground nodes. Rendezvous (air–ground) edges are added only when both landing and ground reachability constraints are satisfied, yielding a sparsely coupled graph. We then formulate viewpoint and tour selection as instances of the Orienteering Problem (OP), enabling each robot to choose an informative and energy-feasible tour within energy budgets. Two OP instances are then coupled through a release–collect mechanism that synchronizes the UAV’s aerial exploration with the UGV’s ground exploration and support. This process operates iteratively: after executing an exploration tour, the map is updated, PRM is expanded to cover newly discovered regions, the UAV is recharged, and exploration continues with new tours.

\subsection{Building the PRM}
\label{subsec:prm}
We construct an incremental PRM that progressively expands the graph as new regions of the environment become available. We use a dual-layer strategy to construct an air-ground coupled roadmap. The roadmap $G{=}(V,E)$ is constructed by sampling the aerial space and generating feasible ground configurations through projections of aerial samples when possible and additional sampled ground nodes. This layered construction embeds rendezvous feasibility directly into the roadmap structure, rather than checking it post hoc during planning.

We extend the vision-based heuristic frontier detection and incremental PRM construction proposed in \cite{xu2023heuristic}, originally designed for single UAV exploration, to our heterogeneous setting (collaborative UAV-UGV team). The approach first detects heuristic frontier regions from the 3D occupancy map: voxel map is sliced into multiple 2D layers with different heights. Then, each of these layers are converted into images where grey pixels are unexplored regions. Blob detection is used to identify heuristic frontier regions as circles at specific heights. These detected heuristic frontiers guide the aerial sampling process. The PRM construction then proceeds with three stages: (i) frontier-guided sampling, (ii) local region sampling near UAV's current position to ensure node density, (iii) and global region sampling for coverage. Sampled nodes are validated for collision-free connectivity and appropriate spacing from existing nodes before being added to the PRM. Sampling is density-aware, terminating after $\mathcal{N}_{\max}^{uav}$ failed attempts. For the details of the heuristic frontiers and sampling, we refer the readers to \cite{xu2023heuristic}.

\begin{figure}[htpb]
    \centering
    \includegraphics[width=1.0\linewidth]{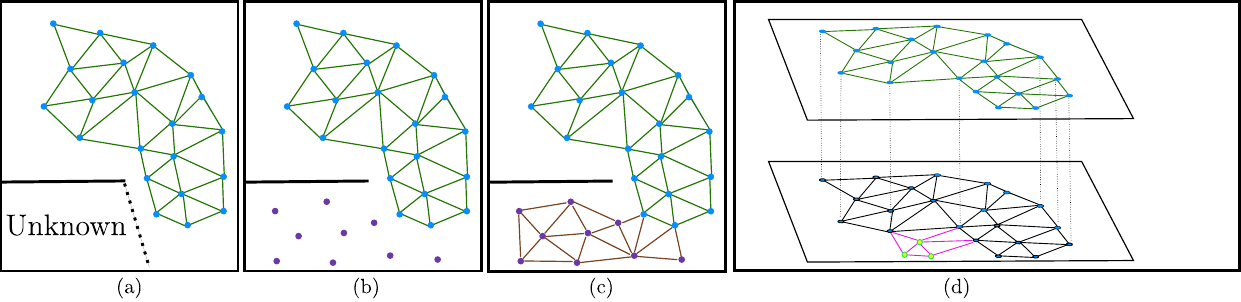}
    \caption{Incremental dual-layer PRM construction. (a-c) Top view of UAV layer construction (purple nodes). (d) 3D view showing ground layer generated through 
    aerial node projection and independent ground sampling (green nodes), interconnected by 
    rendezvous edges (dashed lines).}
    \label{fig:prm}
\end{figure}

To extend this approach to our setting, we introduce ground configurations via two mechanisms. For each valid aerial sample, $\xi^{uav}_{i}$, we compute a projected configuration on ground $\xi^{ugv}_{i}{=}\pi(\xi^{uav}_{i})$. If $\xi^{ugv}_{i}$ is a valid UGV configuration, it is added to the graph, and a corresponding rendezvous edge $(\xi^{uav}_{i},\xi^{ugv}_{i})$ is created. This edge represents a feasible takeoff/landing maneuver. Secondly, we independently sample additional ground configurations to capture terrain regions that may not be reachable with aerial projections alone. We introduce a similar density-aware sampling termination condition for the UGV, $\mathcal{N}_{\max}^{ugv}$. The resulting roadmap can be decomposed into two subgraphs for each robot, interconnected by \textit{rendezvous edges}. The \textit{UAV Subgraph} $G_{uav} {=} (V_{uav}, E_{uav})$ contains feasible flight paths satisfying aforementioned constraints. The \textit{UGV Subgraph} $G_{ugv} {=} (V_{ugv}, E_{ugv})$ contains feasible traversable paths on the terrain. These layers are linked by the set of rendezvous edges $E_{rnd} {=} \{(\xi_i^{uav}, \xi_i^{ugv}) {\in} V_{uav} {\times} V_{ugv} {\mid} \xi_i^{ugv} {=} \pi(\xi_i^{uav})\}$, representing feasible air-ground connections. Then, the complete edge set is $E {=} E_{uav} {\cup} E_{ugv} {\cup} E_{rnd}$. Fig. \ref{fig:prm} demonstrates a representative PRM construction.

\begin{algorithm}[htpb]
\caption{Dual-Layer PRM Generation}
\label{alg:layered-prm}
\renewcommand{\AlCapSty}[1]{\normalfont\footnotesize{\textbf{#1}}\unskip}
\footnotesize
\SetAlgoLined
\DontPrintSemicolon
\SetKwInOut{Input}{Input}
\SetKwInOut{Output}{Output}
\Input{$G$ (Current roadmap),
$\mathcal{M}$ (Current occupancy map),
$\xi_r^{uav}, \xi_r^{ugv}$ (Current robot configurations), 
$d_{\min},\,d_{\max}$ (Distance thresholds), 
$\mathcal{N}_{\max}^{uav},\,\mathcal{N}_{\max}^{ugv}$, (Failure thresholds)}

\Output{$G=(V,E)$}
\textbf{Initialize:} $\mathcal{N}_{\mathrm{fail}}^{uav} \gets 0, \mathcal{N}_{\mathrm{fail}}^{ugv} \gets 0$\;
$\mathcal{S}_f \gets \textsc{GetHeuristicFrontiers}(\mathcal{M})$\;
\While{$\mathcal{N}_{\mathrm{fail}}^{uav} < \mathcal{N}_{\max}^{uav}$}{
  $f_{target} \gets \textsc{WeightedSampling}(\mathcal{S}_f)$\;
  $\mathcal{N} \gets \textsc{kNN}(V \cap \Xi_{uav}, f_{target})$\;
  \ForEach{$\xi_i$ \textbf{in} $\mathcal{N}$}{
    $\xi_{ext} \gets \textsc{extendTowardsFrontier}(\xi_i, f_{target})$\;
    \If{$\mathcal{M}.\textsc{isConfigFree}(\xi_{ext})$}{
      $\xi_{nn} \gets \textsc{NN}(V, \xi_{ext})$\;
      \If{$d_{\min} \le \|\xi_{ext} - \xi_{nn}\|_2 \le d_{\max}$}{
        $V \gets V \cup \{\xi_{ext}\}$; \;
      }
      \Else{
        $\mathcal{N}_{\mathrm{fail}}^{uav} \gets \mathcal{N}_{\mathrm{fail}}^{uav} + 1$\;
      }
    }
  }
}
$V \gets \textsc{localSampling}(V, \mathcal{M}, \xi_r^{uav}, \Xi_{uav})$ \;
$V \gets \textsc{globalSampling}(V, \mathcal{M}, \Xi_{uav})$\;
$E \gets \textsc{connectNewUAVConfigs}(\mathcal{M}, V \cap \Xi_{uav})$\;
\ForEach{$\xi_i^{uav} \in V \cap \Xi_{uav}$}{
  $\xi_i^{ugv} \gets \textsc{ProjectConfig}(\xi_i^{uav})$\;
  \If{\textbf{not} \textsc{Invalid}$(\xi_i^{ugv})$}{
    $V \gets V \cup \{\xi_i^{ugv}\}$; \;
    $E \gets E \cup \{(\xi_i^{uav}, \xi_i^{ugv})\}$\;
  }
}
\While{$\mathcal{N}_{\mathrm{fail}}^{ugv} < \mathcal{N}_{\max}^{ugv}$}{
  $f_{target} \gets \textsc{weightedSampling}(\mathcal{S}_f)$\;
  $\mathcal{N} \gets \textsc{kNN}(V \cap \Xi_{ugv}, f_{target})$\;
  \ForEach{$\xi_i$ \textbf{in} $\mathcal{N}$}{
    $\xi_{ext} \gets \textsc{extendTowardsFrontier}(\xi_i, f_{target})$\;
    \If{$\mathcal{M}.\textsc{isConfigFree}(\xi_{ext})$}{
      $\xi_{nn} \gets \textsc{NN}(V, \xi_{ext})$\;
      \If{$d_{\min} \le \|\xi_{ext} - \xi_{nn}\|_2 \le d_{\max}$}{
        $V \gets V \cup \{\xi_{ext}\}$; \;
      }
      \Else{
        $\mathcal{N}_{\mathrm{fail}}^{ugv} \gets \mathcal{N}_{\mathrm{fail}}^{ugv} + 1$\;
      }
    }
  }
}
$V \gets \textsc{localSampling}(V, \mathcal{M}, \xi_r^{ugv}, \Xi_{ugv})$\;
$V \gets \textsc{globalSampling}(V, \mathcal{M}, \Xi_{ugv})$\;
$E \gets \textsc{connectNewUGVConfigs}(\mathcal{M}, V \cap \Xi_{ugv})$\;
\Return $G=(V,E)$
\end{algorithm}
Algorithm 1 constructs the dual-layer PRM. Line 1 initializes the failure 
counters for both layers. Line 2 gets the heuristic frontier regions from the 
occupancy map. The UAV layer construction (lines 3-16) proceeds in three stages. In the frontier-guided sampling stage (lines 3-13), the algorithm samples a heuristic frontier region (line 4), identifies neighboring aerial nodes to that frontier via a $k$-nearest neighbor (kNN) query (line 5).\footnote{On first iteration, $V$ is empty and kNN at line 5 returns no neighbors, causing the frontier-guided stage to terminate. Sampling proceeds directly to local and global stages (lines 14-15) to bootstrap the initial roadmap.} Each neighbor is extended towards a frontier by a random distance within $[d_{\min}, d_{\max}]$ (line 6-7), which biases the roadmap growth towards unexplored regions. Extension candidates are validated for collisions (line 8) and spacing from existing nodes using a nearest neighbor (NN) lookup (lines 9-10) before being added to the graph (line 11). Nodes failing this check increment the failure count $\mathcal{N}^{uav}_{\text{max}}$. Following frontier-guided sampling, local sampling (line 14) augments $V$ with new nodes drawn near the UAV's current configuration $\xi_r^{\text{uav}}$ to ensure local density, while global sampling (line 15) adds nodes across the entire aerial feasible space $\Xi_{\text{uav}}$ for broader coverage. Line 16 connects the newly added UAV configurations, given that they are collision-free and within distance threshold. For each newly added UAV node, a ground configuration is obtained by projection (lines 17-18). The projected configuration is validated (line 19), rejected if it results in a collision or no feasible takeoff/landing maneuver exist between UAV and UGV configurations, and added to graph along with its rendezvous edge (lines 20-21). Additionally, UGV layer construction (lines 22-35) applies the same three-stage sampling process to the ground with its own failure rate $\mathcal{N}_{\max}^{ugv}$: frontier-guided sampling (lines 22-32), local sampling near the UGV's current configuration $\xi_r^{\text{ugv}}$ and global sampling over $\Xi_{\text{ugv}}$ to augment $V$ with new nodes (lines 33-34). Then, appropriate ground nodes are connected (line 35). Finally, updated roadmap is returned (line 36).

\color{black}
\subsection{Orienteering Problem}
\label{subsec:OP}
The Orienteering Problem (OP) is defined on a graph $G{=}(V,E)$, where each vertex $v_i {\in} V$ is associated with a non-negative reward $S_i$ and each edge $(i,j) \in E$ has a traversal cost $t_{ij}$. Given a distance budget $T_{\max}$, a fixed $v_1$ and a fixed $v_N$ (start and end vertex), the objective is to find a path from $v_1$ to $v_N$. This path must visit a subset of intermediate vertices such that the total collected reward is maximized without exceeding $T_{\max}$. OP is an NP-hard problem requiring approximate solvers and heuristics; standard formulation and solution approaches are detailed in~\cite{vanst2011op}. 

This formulation is naturally suited for energy-aware exploration. The objective  maps directly to maximizing information gain from candidate viewpoints in the roadmap. Crucially, the hard budget constraint provides a mechanism to respect the UAV's maximum flight time $\overline{\tau_a}$ and the UGV's rendezvous window. By solving coupled OPs on the dual-layer graph (i.e., one for UAV, one for UGV), the team can autonomously generate an efficient and energy-feasible tour.

\begin{remark}
Our framework is independent of sensor modality. For omnidirectional sensors (e.g., LiDAR), the information gain is invariant of heading: $S_i {=} \mathcal{G}(\xi_{pos,i})$, where $\mathcal{G}$ is the information gain function. For directional sensors (e.g., RGB-D), the reward is computed as the maximum gain over discretized yaw directions: $S_i {=} \max_{\psi \in \Psi_k} \mathcal{G}(\xi_{pos,i}, \psi)$. 
\end{remark}

\subsection{UAV-UGV Coordination}
\label{subsec:coordination}
We plan the exploration as sequences of coordinated tours. For each tour, we solve two distinct but coupled OPs, one for the UAV and one for the UGV. The coupling between the two OPs arises through the shared release–collect pair, which fixes the start and end vertices of both problems to enforce rendezvous at the start and end of each tour.

The UAV-OP is defined on the weighted subgraph $G_{uav} = (V_{uav}, E_{uav}, w_{uav})$, where $V_{uav} = V \cap \Xi_{uav}$, and $E_{uav} \subseteq E$ are the edges in $G$ connecting nodes in $V_{uav}$, and $w_{uav}$ is the distances between respective UAV positions. The objective is maximizing collected UAV rewards. The rewards $S_i^{uav}$ are set to the UAV sensor information gain for all $v_i \in V_{uav}$. The distance budget is set to the battery endurance: $\overline{\tau_a} \cdot v_{uav}$, where $v_{uav}$ is average UAV speed.

The UGV-OP is defined on the weighted subgraph $G_{ugv} = (V_{ugv}, E_{ugv}, w_{ugv})$, where $V_{ugv} = V \cap \Xi_{ugv}$ and $E_{ugv} \subseteq E$ are the edges in $G$ connecting nodes in $V_{ugv}$, and $w_{ugv}$ is the distances between respective UGV positions. Its objective is to maximize its own collected rewards $S_i^{ugv}$ (set to UGV sensor information gain for $v_i \in V_{ugv}$) along its path. The distance budget is set to $\overline{\tau_a} {\cdot} v_{ugv}$, where $v_{ugv}$ is average UGV speed, serves as a rendezvous constraint. It ensures that the UGV rendezvous with the UAV within $\overline{\tau_a}$. 

The two OPs are fundamentally coupled by the selection of the \textit{collect} pair $(v_{col}, v'_{col})$, where $v_{col} \in V_{ugv}$ is the physical \textit{collect} location and $v'_{col} \in V_{uav}$ is its aerial projection. The \textit{release} pair, $(v_{rel}, v'_{rel})$, is: the initial deployment configuration for the first tour, and for subsequent tours, it is inherited from the \textit{collect} of the preceding tour. Since the rendezvous (i.e., \textit{release} and \textit{collect}) edges for takeoff and landing are determined in advance, we treat these costs as constants.

The selection of the \textit{collect} pair $(v_{col}, v'_{col})$ is the critical decision ensuring mission-level feasibility. An arbitrary choice for this pair could lead to a scenario where the UAV cannot reach within $\overline{\tau_a}$ or UGV cannot traverse. Therefore, a pair should be selected such that $(v'_{col}, v_{col})$ is reachable by both robots within $\overline{\tau_a}$.

We employ a utility-driven selection strategy to identify this pair. We extract a set of candidate aerial viewpoints $V_{\text{cand}} \subseteq V_{\text{uav}}$ with strictly positive information gain, then rank them by exploration utility. For each candidate $v'_{\text{cand}} \in V_{\text{cand}}$, we get the corresponding projection $v_{\text{cand}} \in V_{\text{ugv}}$. Prior to solving the OPs, we score each candidate pair by a utility balancing the aerial info gain against the ground travel cost:
\begin{equation}
\label{eq:utility_function}
    U(v_{cand}, v'_{cand}) = S^{uav}(v'_{cand}, \mathcal{M}) \cdot \exp(-\lambda \cdot \Delta^{ugv}_{path}),
\end{equation}
\color{black}

where $\Delta^{ugv}_{path}$ is the shortest-path distance on $G_{ugv}$ from 
$v_{rel}$ to $v_{cand}$. Here, $S^{uav}(v'_{cand}, \mathcal{M})$ is the UAV information gain at the candidate viewpoint which represents the unexplored volume observable from $(v'_{cand})$ in current $\mathcal{M}$. The multiplicative structure of \eqref{eq:utility_function} discounts high-info-gain but distant candidates, trading off exploration gain with UGV repositioning cost. This tradeoff is controlled by $\lambda$.\footnote{$\lambda$ can be tuned based on the UAV-UGV speed ratio; higher values are suitable when the UGV is significantly slower than the UAV, penalizing long UGV travels. $\lambda = 0$ recovers pure aerial information gain selection.} Once the best candidate is selected, we set $v_{col} \leftarrow v_{cand}$ and $v'_{col} \leftarrow v'_{cand}$, allowing both agents to generate their trajectories on their respective subgraphs.

\begin{algorithm}[htpb]
\caption{Coupled Exploration Tour Generation}
\label{alg:rendezvous-selection}
\renewcommand{\AlCapSty}[1]{\normalfont\footnotesize{\textbf{#1}}\unskip}
\footnotesize
\SetAlgoLined
\DontPrintSemicolon
\SetKwInOut{Input}{Input}
\SetKwInOut{Output}{Output}
\Input{$G_{uav}$ (UAV weighted subgraph), $G_{ugv}$ (UGV weighted subgraph), $(v_{rel}, v'_{rel})$ (Release pair), $V_{\text{cand}}$ (Candidate viewpoints), $\overline{\tau_a}$ (Max flight time), $v_{uav}, v_{ugv}$ (UAV-UGV average speeds), $\mathcal{M}$ (Map), $\lambda$ (Penalty parameter)}
\Output{$\sigma_{uav}, \sigma_{ugv}$ (UAV and UGV exploration paths)}
\DontPrintSemicolon

\textbf{Initialize: }$L \leftarrow \emptyset$\;

\ForEach{$v'_{cand} \in V_{\text{cand}}$}{
    $v_{cand} \leftarrow \textsc{GetProjection}(v'_{cand})$\;
    
    $\Delta^{uav}_{path} \leftarrow \textsc{Dijkstra}(G_{uav}, v'_{rel}, v'_{cand})$\;
    $\Delta^{ugv}_{path} \leftarrow \textsc{Dijkstra}(G_{ugv}, v_{rel}, v_{cand})$\;

    \If{$\Delta^{uav}_{path} \le\overline{\tau_a} \cdot v_{uav}$ \textbf{and} $\Delta^{ugv}_{path} \le \overline{\tau_a} \cdot v_{ugv}$}{
        $R \leftarrow \textsc{GetInfoGain}(v'_{cand}, \mathcal{M})$\;
        $U \leftarrow R \cdot \exp(-\lambda \cdot \Delta^{ugv}_{path})$\;
        $L.\text{add}((v_{cand}, v'_{cand}, U))$\;
    }
}

\If{$L$ \textbf{is not} empty}{
    $(v_{col}, v'_{col}, U_{col}) \leftarrow \arg\max_{(v, v', U)\in L} U$\;
    $\sigma_{ugv} \leftarrow \textsc{UGV-OP}(G_{ugv}, v_{rel}, v_{col}, \overline{\tau_a} \cdot v_{ugv})$\;
    $\sigma_{uav} \leftarrow \textsc{UAV-OP}(G_{uav}, v'_{rel}, v'_{col}, \overline{\tau_a} \cdot v_{uav})$\;
}

\Else{
    \eIf{$\textsc{MaxInfoGain}(V_{uav}\cup V_{ugv},\mathcal{M}) > 0$}{
        $v_{col} \leftarrow \textsc{MaxInfoGainNodeProjection}(V_{uav}\cup V_{ugv}, \mathcal{M})$\;
    }{
        $v_{col} \leftarrow v_{rel}$\;
    }
    $\sigma_{ugv} \leftarrow \textsc{Dijkstra}(G_{ugv}, v_{rel}, v_{col})$\;
    $\sigma_{uav} \leftarrow \emptyset$\footnotemark\;
}

\Return{$\sigma_{uav}, \sigma_{ugv}$}\;
\end{algorithm}

\footnotetext{As a slight abuse of notation we use $\sigma_{uav} {=} \emptyset$ to denote the UAV does not move.}

\color{black}

Algorithm \ref{alg:rendezvous-selection} handles selecting the best \textit{collect} pair and generating the coordinated tours for the UAV-UGV team. Line 1 initializes an empty list $L$ to store feasible candidates. The algorithm then iterates through the set of high-info-gain aerial candidates $V_{\text{cand}}$. For each aerial candidate $v'_{cand}$, Line 3 gets the ground projection $v_{cand}$. Lines 4-5 compute the travel costs for each agent independently. Line 6 checks feasibility: the candidate pair is valid only if the UAV can reach it within distance $\overline{\tau_a} \cdot v_{uav}$ and the UGV can reach it within distance $\overline{\tau_a} \cdot v_{ugv}$. If valid, the candidate's utility is computed in Lines 7-9 using (\ref{eq:utility_function}) and added to $L$. Lines 11-13 handle the tour generation. If $L$ is not empty (i.e., feasible candidates exist), the algorithm selects the pair $(v_{col}, v'_{col})$ with the highest utility. Then, OP instances are solved in lines 12-13 (since OP is NP-hard \cite{vanst2011op}, we use a heuristic approach to get a tractable solution \cite{campos2014op}) UGV-OP finds a path on $G_{ugv}$ to maximize ground rewards (info gain), while the UAV-OP finds a path on $G_{uav}$ to maximize aerial rewards (info gain), both constrained to rendezvous with each other at collect location. Lines 14-20 handle the fallback.\footnote{The fallback is triggered when no candidate pair is reachable by both robots within $\overline{\tau_a}$, for example when the UAV has exhausted all reachable viewpoints from the current team position. Since no aerial tour is executed, no rendezvous is required, UGV can re-position freely in the fallback.} The algorithm checks whether any node in $V_{uav} \cup V_{ugv}$ yields positive information gain (line 15). The UGV relocates towards the maximum-gain node: if it belongs to $V_{ugv}$, UGV navigates directly; if it belongs to $V_{uav}$, the UGV navigates to the ground node closest to its ground projection (line 16). Otherwise, the UGV stays put (line 18). In all fallback cases, the UAV remains on the UGV (line 20). Finally, exploration paths, $\sigma_{uav}, \sigma_{ugv}$ are returned (line 21).

\subsection{Exploration Pipeline}
We integrate the methods described in \ref{subsec:prm}, \ref{subsec:OP}, and \ref{subsec:coordination} into a unified exploration pipeline. This framework orchestrates the UAV-UGV team to execute a sequence of energy-constrained tours until an exploration criterion is met.

\begin{figure}[htpb]
    \centering
    \includegraphics[width=0.8\linewidth]{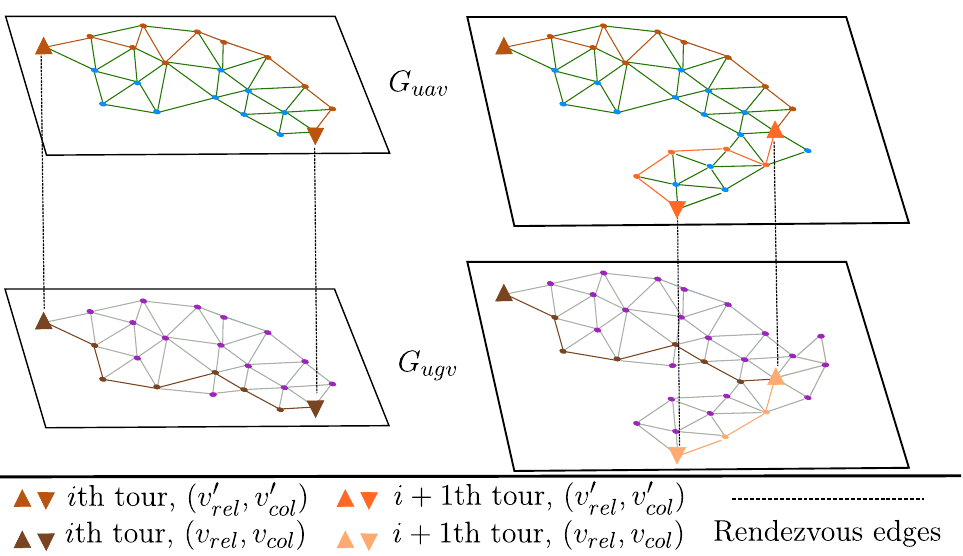}
    \caption{Consecutive exploration tours of the UAV and the UGV.}
    \label{fig:op}
\end{figure}

The core exploration cycle consists of three distinct phases. Initially, in \textit{construction phase}, with Alg.\ref{alg:layered-prm}, $G$ is incrementally expanded to capture new regions, and information gains $S^{uav}_i, S^{ugv}_i$ are updated. Secondly, in \textit{planning phase}, Alg.\ref{alg:rendezvous-selection} generates $(\sigma_{uav}, \sigma_{ugv})$ that maximize information gain within respective energy and rendezvous limits. Finally, in \textit{execution phase}, the team executes their respective paths, and updates the occupancy map. \textit{Collect} pair of each tour serves as the release pair for the next tour. Fig.\ref{fig:op} demonstrates the paths of two consecutive exploration tours for both robots.

\begin{algorithm}[htpb]
\caption{Integrated Exploration Pipeline}
\label{alg:pipeline}
\renewcommand{\AlCapSty}[1]{\normalfont\footnotesize{\textbf{#1}}\unskip}
\footnotesize
\SetAlgoLined
\DontPrintSemicolon
\SetKwInOut{Input}{Input}
\SetKwInOut{Output}{Output}
\Input{$\mathcal{M}_0$ (Initial map), $v_{rel}, v'_{rel}$ (Release pair), $d_{\min}, d_{\max}$ (Distance thresholds), $\mathcal{N}_{\max}^{uav}, \mathcal{N}_{\max}^{ugv}$ (Failure thresholds), $\overline{\tau_a}$ (Max flight time), $v_{uav}, v_{ugv}$ (UAV-UGV average speeds), $\lambda$ (Penalty parameter)}
\Output{$\mathcal{M}$ (Final map)}
\textbf{Initialize:} $\mathcal{M} \gets \mathcal{M}_0,\; G \gets \emptyset$\;
\While{exploration criterion\footnotemark not satisfied}{
  $G\gets \textsc{Alg.1}(G, \mathcal{M}, v_{rel}, v'_{rel}, d_{\min}, d_{\max}, \mathcal{N}_{\max}^{uav}, \mathcal{N}_{\max}^{ugv})$\;
  \textsc{UpdateInfoGain}$(G, \mathcal{M})$\;
  $V_{cand} \gets \textsc{GetCandidates}(V_{uav})$\;
  $(\sigma_{uav}, \sigma_{ugv}) \gets$
  \hspace{1em}$\textsc{Alg.2}(G_{uav}, G_{ugv}, v_{rel}, v'_{rel}, V_{cand}, \overline{\tau_a}, v_{uav}, v_{ugv}, \mathcal{M}, \lambda$\;
  \textsc{Execute}$(\sigma_{uav}, \sigma_{ugv})$\;
  $\mathcal{M} \gets \textsc{UpdateMap}(\mathcal{M})$\;
  \textsc{RechargeUAV}\;
  $v_{rel} \gets \textsc{End}(\sigma_{ugv})$\;
  \eIf{$\sigma_{uav} \neq \emptyset$}{
    $v'_{rel} \gets \textsc{End}(\sigma_{uav})$\;
  }{
    $v'_{rel} \gets \textsc{GetAerialProjection}(v_{rel})$\;
  }
}
\Return $\mathcal{M}$
\end{algorithm}
\footnotetext{In practice, the exploration criterion can be defined as a coverage threshold or a time limit, e.g., 95\% exploration or 30 minutes.}

Algorithm \ref{alg:pipeline} details the pipeline. In line 1, the map $\mathcal{M}$ and the roadmap $G$ are initialized. The core loop (lines 2-14) continues until exploration criterion is satisfied. In each iteration, the roadmap is extended using Algorithm \ref{alg:layered-prm} and info gain are updated to reflect the current map state (lines 3-4). Then, high-utility candidates are identified and Algorithm \ref{alg:rendezvous-selection} is used to generate  trajectories $(\sigma_{uav}, \sigma_{ugv})$ (lines 5-6). Finally, the robots execute these paths (line 7), and the map is updated with new sensor data (line 8).\footnote{Map updates can occur at a configurable frequency depending on computational resources, we present a single update per tour as an example.} The UAV then charges on the UGV (line 9), ensuring it is fully charged before the next tour. Then, $v_{rel}$ is set to the UGV's end position (line 10), and $v'_{rel}$ is set to the UAV's end position if it flew, or its projection otherwise (lines 11-14). Finally, explored map $\mathcal{M}$ is returned (line 15).

\begin{remark}[Feasibility and Exploration Progress]
By construction, each UAV–UGV tour generated by Alg.~\ref{alg:pipeline} is planned under an explicit UAV energy budget and executed only if the selected release–collect pair guarantees feasible rendezvous within the UAV's maximum flight time $\overline{\tau_a}$. Since tours are constructed over a PRM with collision-validated edges, every executed tour is collision-free with respect to the static environment model. Furthermore, unless both robots remain stationary, every tour yields positive information gain, ensuring $\mathcal{V}_{\text{mapped}}$ increases.
\end{remark}

\color{black}
\section{Simulations}
\subsection{Setup}
The simulation architecture was built in ROS with C++, using Gazebo on a PC equipped with i9-14900K, 64GB RAM, and RTX4090. We build upon the open-source framework of \cite{xu2023heuristic}, which provides the UAV model (with an RGB-D camera), occupancy mapping, and incremental PRM construction, and extend it to our heterogeneous UAV-UGV setting as described in Section~\ref{sec:proposed_method}. For the UGV model, Scout Mini (with a 3D LiDAR) was used.

\begin{figure}[htpb]
    \centering
    \includegraphics[width=1.0\linewidth]{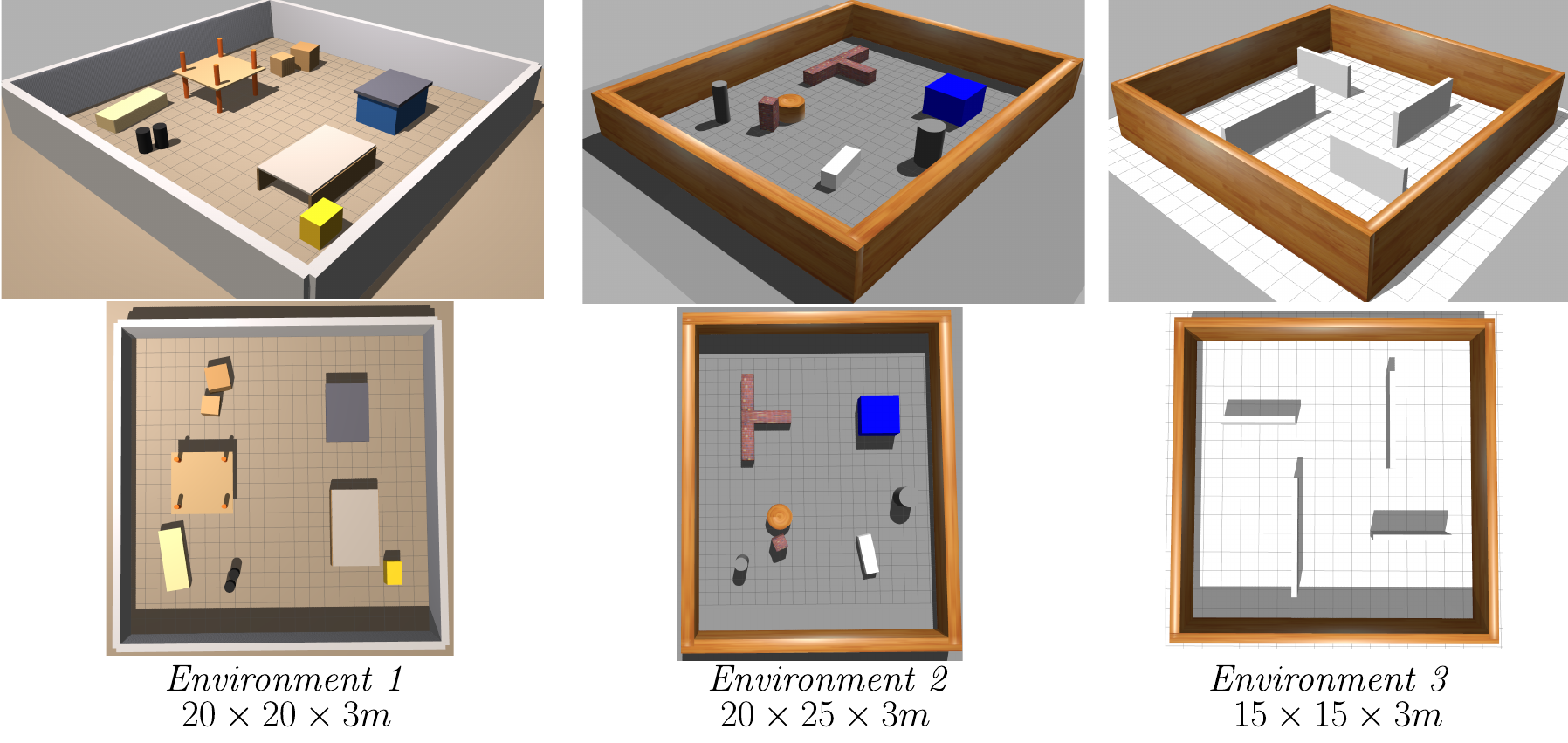}
    \caption{Environments used in the simulations.}
    \label{fig:environments}
\end{figure}

We use three environments of different sizes and complexity, as in Fig. \ref{fig:environments}. These environments are designed to utilize the complementary capabilities of the UAV-UGV team. For instance, \textit{Environment 1} includes a tunnel inaccessible to the UAV and elevated scaffolding unreachable by the UGV, making neither robot alone sufficient for full coverage. Across all environments, cluttered layouts with passages, open regions, and occlusion-causing obstacles capture a range of challenges that motivate collaborative exploration.

Both robots start co-located at the origin. To obtain $\mathcal{V}_{\text{mapped}}^0$ (initially mapped space), the UAV takes off and performs a 360 degrees scan at the beginning of each mission. Sensor and algorithm parameters are summarized in Table~\ref{tab:sim_params}. 

UAV's max flight time $\overline{\tau_a} {=} 120s$ and average speed $v_{uav} {=} 0.25 m/s$ yield an effective UAV-OP distance budget of $30m$,  while $v_{ugv} {=} 0.17 m/s$ yields an effective UGV-OP distance budget of approximately $20m$ (see Subsec. \ref{subsec:OP}). The average speeds are conservative, capturing the effective speed over a full tour: at each visited viewpoint, the UAV decelerates, rotates to observe the viewpoint, and reorients toward the next waypoint. Similarly, the UGV does frequent heading adjustments to align with its planned path.

\begin{table}[htpb]
\centering
\caption{Simulation Parameters}
\label{tab:sim_params}
\resizebox{\columnwidth}{!}{%
\begin{tabular}{l c l c}
\hline
\textbf{Parameter} & \textbf{Value} & \textbf{Parameter} & \textbf{Value} \\
\hline
$r$ (voxel resolution) & $0.1m$ 
& UAV sensing range & $2.0m$ \\
UAV horizontal FOV & $90^{\circ}$ 
& UAV vertical FOV & $90^{\circ}$ \\
UGV sensing range & $5.0m$ 
& UGV horizontal FOV & $360^{\circ}$ \\
UGV vertical FOV & $30^{\circ}$ 
& $\overline{\tau_a}$ (Max flight time) & $120s$ \\
$v_{uav}$ (UAV avg. speed) & $0.25\ m/s$ 
& $v_{ugv}$ (UGV avg. speed) & $0.17\ m/s$ \\
$\tau_c$ (charging duration) & $5.0s$ 
& $\lambda$ (utility weight) & $0.001$ \\
$d_{\min}$ (min node spacing) & $1.2m$ 
& $d_{\max}$ (max node spacing) & $3.0m$ \\
$\mathcal{N}_{\max}^{uav}$ (UAV sampling limit) & $100$ 
& $\mathcal{N}_{\max}^{ugv}$ (UGV sampling limit) & $100$ \\
\hline
\end{tabular}
}
\end{table}

Each mission runs until 95\% exploration is reached or 30 minutes pass, with results averaged over 5 runs per environment with standard deviations reported.\footnote{Plots report total mission time (including exploration, computation and recharging times), while tables report active exploration time (only exploration, excluding computation and recharging times).}

\subsection{Results}
\begin{figure}[htpb]
    \centering
    \includegraphics[width=1.0\linewidth]{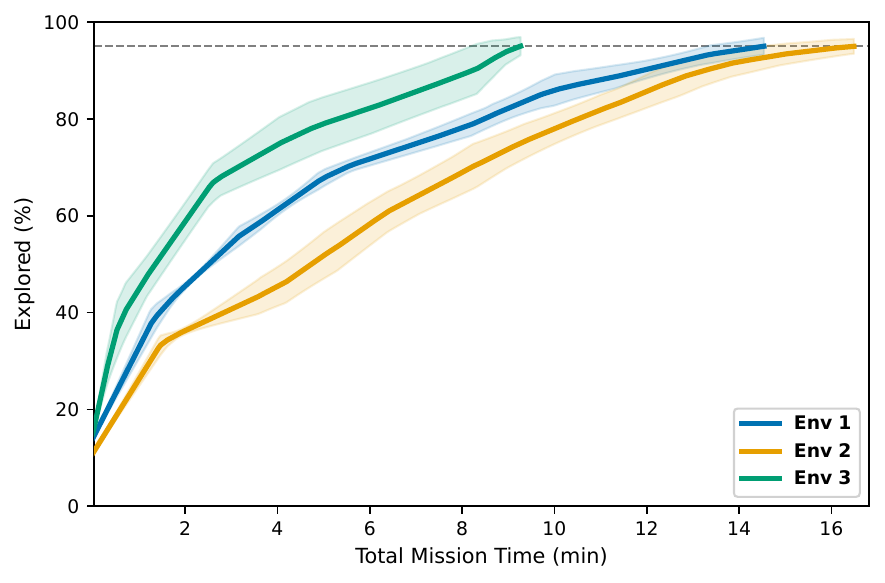}
    \caption{Exploration progress across all environments. Bold lines show the mean and shaded regions indicate $\pm$1 standard deviation. The x-axis represents total mission time, including, computation and recharging time.}
    \label{fig:coverage_combined}
\end{figure}

Fig.~\ref{fig:coverage_combined} shows our method's exploration progress over mission time. All environments exhibit rapid initial exploration growth. Exploration is completed the fastest in \textit{Environment 3}. Whereas, \textit{Environments 1} and \textit{2} need longer times due to their larger sizes and more complex layouts. Notably, in all runs, the UAV never exceeded $\overline{\tau_a}$.

Table~\ref{tab:sim_results} demonstrates further metrics. Total computation time remains significantly small across all environments despite the larger number of tours required by \textit{Environments 1} and \textit{2}, demonstrating the computational performance.
\begin{table}[htpb]
\centering
\caption{Simulation Results per Environment. $t_{\mathrm{exp}}^{80\%}$ and $t_{\mathrm{exp}}^{95\%}$: active exploration time to reach 80\% and 95\% exploration. $t_{\mathrm{comp}}^{95\%}$: total computation time until 95\% exploration. $N_{\mathrm{tours}}$: number of tours until termination.}\label{tab:sim_results}
\resizebox{\columnwidth}{!}{%
\begin{tabular}{lcccccc}
\toprule
\textbf{Env} 
& \textbf{$t_{\mathrm{exp}}^{80\%}$ (min)} 
& \textbf{$t_{\mathrm{exp}}^{95\%}$ (min)} 
& \textbf{$t_{\mathrm{comp}}^{95\%}$ (s)} 
& \textbf{$N_{tours}$} \\
\midrule
1 & $8.12 \pm 0.60$ & $14.13 \pm 1.28$ & $3.65 \pm 0.63$ & $8.0 \pm 0.7$ \\
2 & $9.97 \pm 0.87$ & $15.46 \pm 1.90$ & $4.70 \pm 1.10$ & $9.2 \pm 0.8$ \\
3 & $5.23 \pm 1.52$ & $8.63 \pm 0.65$  & $1.86 \pm 0.21$ & $5.0 \pm 0$ \\
\bottomrule
\end{tabular}%
}
\end{table}

\begin{figure}[htpb]
    \centering
    \includegraphics[width=1.0\linewidth]{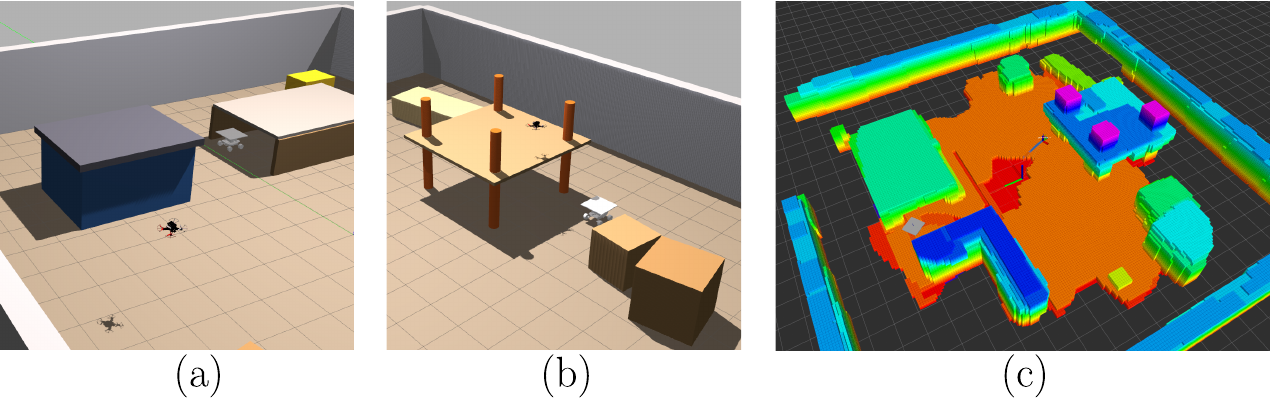}
    \caption{Exploration in \textit{Environment 1}. (a) UGV exploring a tunnel inaccessible to the UAV, (b) UAV exploring an elevated scaffolding unreachable by the UGV, (c) a snapshot of the exploration when the UGV enters the tunnel}
    \label{fig:tunnel}
\end{figure}

We further analyze \textit{Env. 1}, illustrated in Fig.\ref{fig:tunnel}. In this environment, UGV can enter and explore a tunnel that UAV cannot access (Fig.\ref{fig:tunnel}a). Additionally, UAV can explore elevated scaffolding that UGV cannot explore (Fig.\ref{fig:tunnel}b). Neither alone would achieve full coverage in this scenario. 

\subsubsection{Benchmark with Single-UAV}
We compare our method against a single-UAV ablation of our own framework, with the UGV removed. In the baseline, the UAV has the same effective distance budget $\overline{\tau_a} \cdot v_{uav} = 30m$ but must return to a static charging station at the origin after each tour. This isolates the contribution of the UGV as an active co-explorer and mobile charging station.

\begin{figure*}[!t]
    \centering
    \includegraphics[width=1.0\linewidth]{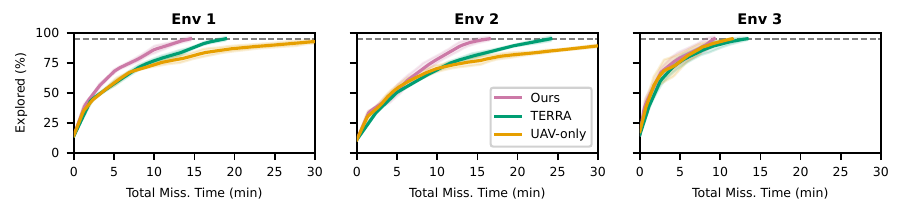}
    \caption{Exploration progress of Our Method vs TERRA vs UAV-only baseline. Bold lines are the means and shaded regions are $\pm$1 standard deviation.}
    \label{fig:benchmark_all}
\end{figure*}

Fig. \ref{fig:benchmark_all} summarizes the exploration progress and mission time. The UAV-only baseline fails to achieve 95\% exploration within 30 minutes in \textit{Environments 1} and \textit{2}. In \textit{Environment 3}, both methods reach 95\% exploration, with our method being approximately 20\% faster. 

\begin{table*}[t!]
\centering
\caption{Comparison: Our Method vs.\ TERRA vs.\ UAV-only. $t_{\mathrm{exp}}^{80\%}$ and $t_{\mathrm{exp}}^{95\%}$: active exploration time to reach 80\% and 95\% exploration. $N_{\mathrm{tours}}$: number of tours until termination. Exp@30 (\%): the percentage of the environment explored by the 30-minute cutoff.}
\label{tab:unified_benchmark}
\resizebox{\linewidth}{!}{%
\begin{tabular}{l ccc ccc cccc}
\toprule
& \multicolumn{3}{c}{\textbf{Ours}}
& \multicolumn{3}{c}{\textbf{TERRA}}
& \multicolumn{4}{c}{\textbf{UAV-only}} \\
\cmidrule(lr){2-4} \cmidrule(lr){5-7} \cmidrule(lr){8-11}
\textbf{Env}
& $t_{\mathrm{exp}}^{80\%}$ (min)
& $t_{\mathrm{exp}}^{95\%}$ (min)
& \textbf{$N_{\mathrm{tours}}$}
& $t_{\mathrm{exp}}^{80\%}$ (min)
& $t_{\mathrm{exp}}^{95\%}$ (min)
& \textbf{$N_{\mathrm{tours}}$}
& $t_{\mathrm{exp}}^{80\%}$ (min)
& $t_{\mathrm{exp}}^{95\%}$ (min)
& {Exp@30 (\%)}
& \textbf{$N_{\mathrm{tours}}$} \\
\midrule
1 & $\mathbf{8.12 \pm 0.60}$ & $\mathbf{14.13 \pm 1.28}$ & $8.0 \pm 0.7$
  & $11.42 \pm 1.19$ & $18.53 \pm 1.22$ & $\mathbf{7.2 \pm 0.4}$
  & $13.65 \pm 1.95$ & $\text{—}$ & $92.8 \pm 1.8$ & $19.5 \pm 1.5$ \\

2 & $\mathbf{9.97 \pm 0.87}$ & $\mathbf{15.46 \pm 1.90}$ & $9.2 \pm 0.8$
  & $13.85 \pm 0.99$ & $24.18 \pm 2.68$ & $\mathbf{8.5 \pm 0.5}$
  & $16.04 \pm 2.28$ & $\text{—}$ & $90.3 \pm 1.5$ & $20.2 \pm 0.8$ \\

3 & $\mathbf{5.23 \pm 1.52}$ & $\mathbf{8.63  \pm 0.65}$ & $\mathbf{5.0 \pm 0.0}$
  & $5.89  \pm 0.91$ & $12.24 \pm 1.33$ & $5.2 \pm 1.1$
  & $5.25 \pm 1.32$ & $10.86 \pm 0.42$ & $\text{—}$ & $9.0 \pm 1.0$ \\
\bottomrule
\end{tabular}}
\end{table*}

Table \ref{tab:unified_benchmark} reports active exploration performance (see Table \ref{tab:sim_results} for metrics details). In \textit{Environments 1} and \textit{2}, the baseline can only reach 92.8\% and 90.3\% exploration before the mission terminates. The number of tours by UAV-only method is significantly larger compared to our method, as the energy-aware UAV must repeatedly visit the static charging station. Because of this, the UAV uses its flight time mainly in transit back to origin, rather than further exploration, leading to diminishing progress. The only environment where the UAV-only baseline achieves 95\% exploration is \textit{Environment 3}, and it takes nearly twice of the tours than our method.

The results show that using a UGV not only enables ground support but also significantly improves exploration efficiency. By repositioning the UGV closer to active exploration regions, the UAV spends less of its flight time on transit and more on exploration.

\color{black}
\subsubsection{Benchmark with UAV-UGV Team}
We adapt \cite{ropero2019terra} (TERRA), a collaborative, energy-aware approach where the UGV transports and recharges the UAV across an exploration area. In TERRA, the UAV executes a series of energy-bounded tours, each taking-off and landing to the stationary UGV. Such structure allows to use OP to plan each tour: the OP budget constraint generalizes TERRA's per-tour energy-bound. While TERRA was originally designed for the case of known target points, its coordination structure translates naturally to an incremental exploration settting.

In this baseline, at each tour: the UGV selects its next (\textit{release}) point via \eqref{eq:utility_function}\footnote{As there is no release-collect mechanism, \eqref{eq:utility_function} is evaluated over release candidates with $\lambda {=} 0$. The UGV has no budget constraint or travel penalty.},  travels there, releases the UAV 
for an energy-bounded exploration tour (with the same distance budget $\overline{\tau_a} \cdot v_{uav} = 30m$), and waits until the UAV returns. The map $\mathcal{M}$ and roadmap  $G$ are updated after each tour. Crucially, the two robots operate \emph{sequentially} rather than concurrently: the UGV does not contribute to exploration while the UAV is airborne for aerial exploration. The UGV only explores while traveling between different \textit{release} points.

Fig.\ref{fig:benchmark_all} demonstrates exploration progress over mission time. Our method reaches 95\% exploration faster than TERRA across all environments. Achieving approximately 23\%, 35\%, and 28\% faster exploration in \textit{Environments 1, 2,} and \textit{3}, respectively.


Table \ref{tab:unified_benchmark} reports active exploration times (see Table \ref{tab:sim_results} for metrics details). TERRA requires fewer tours in \textit{Environment 1} and \textit{2} as the UGV travels freely between release points to release the UAV. Despite this relaxed structure, our method outperforms TERRA across all environments and metrics. The performance gap is most evident in \textit{Environment 2}, where the larger and complex workspace amplifies the downsides of sequential operation. 

The results confirm that the performance gap stems from concurrent exploration: our method extracts useful coverage from both robots simultaneously, whereas TERRA leaves the UGV idle during the most information-rich phases of tours.

\color{black}
\section{Experiments}
To validate our approach, we conducted an experiment with a Crazyflie 2.1+ (UAV) and a TurtleBot 4 (UGV). An OptiTrack motion capture system (covering an area of $6m{\times}6m{\times}3m$) provided localization. Both robots were equipped with 2D sensors: the UAV had a multi-ranger deck serving as a planar LiDAR, while the UGV had RPLIDAR A1M8. The setup was built in ROS2. UGV navigated with a PID controller, while the UAV was controlled with commands for navigation, takeoff, and landing. For UAV, a Kalman filter was used for state estimation, and a Mellinger controller was used for flight control. The environment consists of irregular, tree-like obstacles causing occlusions, making neither aerial nor ground sensing alone sufficient.

\begin{figure}[htpb]
    \centering
    \includegraphics[width=1.0\linewidth]{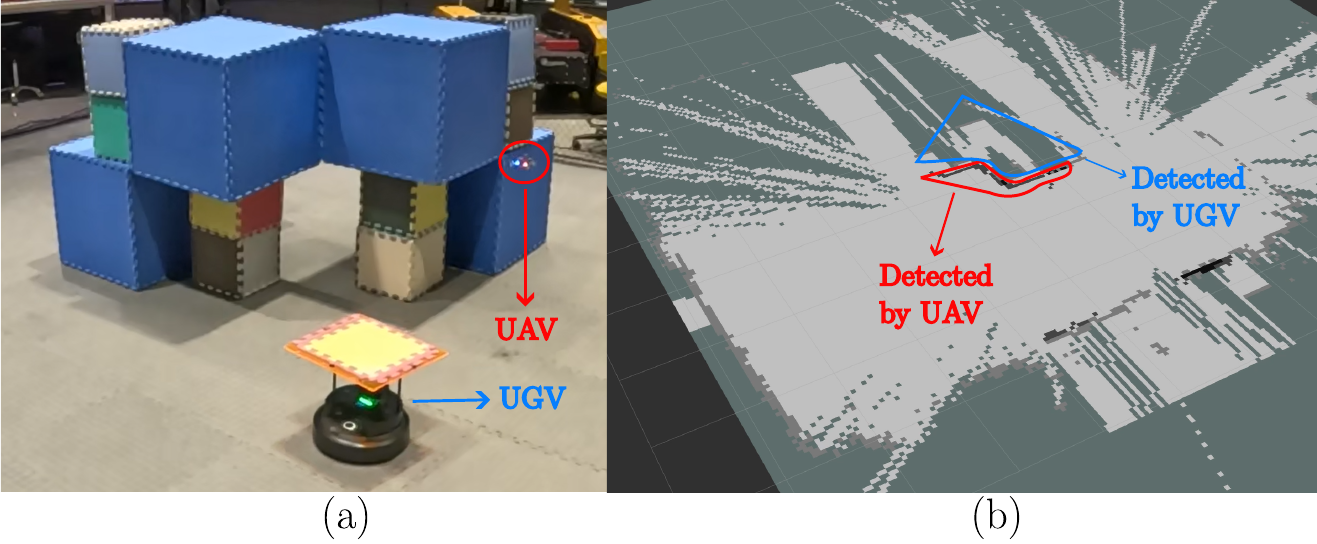}
    \caption{Experiment scenario. (a) A snapshot from the experiment during the first tour, (b) 2.5D Occupancy map built during the first tour of exploration, where light gray corresponds to occupied cells detected by UGV and dark gray corresponds to occupied cells detected by UAV.}
    \label{fig:experiment}
\end{figure}

We used $\overline{\tau_a} {=} 45$s and a conservative $v_{uav} {=} 0.08$ m/s, yielding an effective UAV distance budget of $\overline{\tau_a} \cdot v_{uav} = 3.6$m. This conservative average speed was selected to account for the low flight speed and the dwell of approximately 8 seconds at each visited node, both of which were needed to acquire reliable measurements from the multi-ranger deck. The UGV distance budget was set to $\overline{\tau_a} \cdot v_{ugv} = 3.3$m, yielding $v_{ugv} {\approx} 0.07$ m/s . After each tour, the UAV remained on the UGV for $\tau_c {=} 15$s to model recharging. In total, 4 tours were executed and the team was able to explore more than 95\% of the environment while respecting energy and rendezvous constraints i.e.,  UAV never exceeded $\overline{\tau_a}$. Fig.~\ref{fig:experiment} shows a snapshot of the experiment during the first tour.
\color{black}

\section{Conclusion}

We presented an energy-aware, collaborative exploration framework that uses both the UAV and UGV as active co-explorers under a shared time budget. By sparsely coupling aerial and ground exploration through a density-aware, layered PRM and planning via coordinated Orienteering Problems, we allow both robots co-explore incrementally while ensuring they rendezvous as needed for recharging. Simulations show that the team achieves effective exploration in challenging environments where neither robot alone would suffice. Benchmark comparisons against a UAV-only baseline and a collaborative UAV-UGV baseline demonstrate improvements in exploration performance, achieving significantly faster exploration. Real-world experiments further validate the applicability of our approach.

Future work includes extending our framework to dynamic environments, scaling to multi-UAV-UGV teams, and incorporating learning-based methods to further improve exploration efficiency.

\bibliographystyle{ieeetr}  
\bibliography{references}

\end{document}